# Dust concentration vision measurement based on moment of inertia in gray level-rank co-occurrence matrix

Zhiwen Luo[1], GuohuiLi[1*], Junfeng Du[1], and Jieping Wu[1]

## Abstract

To improve the accuracy of existing dust concentration measurements, a dust concentration measurement based on Moment of inertia in Gray level-Rank Co-occurrence Matrix (GRCM), which is from the dust image sample measured by a machine vision system is proposed in this paper. Firstly, a Polynomial computational model between dust Concentration and Moment of inertia (PCM) is established by experimental methods and fitting methods. Then computing methods for GRCM and its Moment of inertia are constructed by theoretical and mathematical analysis methods. And then developing an on-line dust concentration vision measurement experimental system, the cement dust concentration measurement in a cement production workshop is taken as a practice example with the system and the PCM measurement. The results show that measurement error is within ±9%, and the measurement range is 0.5-1000 mg/m$^3$. Finally, comparing with the filter membrane weighing measurement, light scattering measurement and laser measurement, the proposed PCM measurement has advantages on error and cost, which can be provided a valuable reference for the dust concentration vision measurements.

## 1. Introduction

It's much possible to get pneumoconiosis for some operators who work in the dust environment for a long time, and the life and accuracy of precision instruments are lower than normal when dust is attached to precision instruments. What is more serious is that explosion accident maybe occur when dust concentration reaches the explosion limit, which is a threat to personal safety and property damage [1-2]. The dust concentration measurement is one of key links in controlling the dust pollution; therefore it has necessity and practicability.

Sampling and non-sampling measurement are applied to measure dust concentration. The sampling measurement is always used in some occasions where the real-time measurement is not highly required, such as Membrane Weighing measurement [3], β-ray measurement [4], etc., The Membrane Weighing measurement is often used to calibration because of the advantage of the accuracy [5], and whose measuring error is within 10%. In the non-sampling measurement, the dust concentration is characterized by indirect physical quantity and it can be measured in real-time, such as Charge Induction measurement [6-7], Light Scattering measurement [8-9], Laser measurement [10-11], Vision measurement [13-15], etc., Among them, the measurement accuracy of Charge Induction measurement is affected by the wind speed and self-electrical charges of dust particles [16]. The measurement accuracy of Light Scattering measurement is disturbed by operating current, simultaneously, the operation of them is always complicated [12]. About Laser measurement, it has some limitations, such as point to point measurement mode, difficult calibration and high-cost [11]. As for the Vision measurement, it is widely applied in the measurement field of coal dust [2, 17], chemical powder [13] and furnace dust measurement [15], etc., owing to the advantage on visualization, high-accuracy, face-scanning-measuring mode, etc., which has wide application prospect, but it still has room to improve in measurement accuracy.

* Corresponding author: guohuili@sicnu.edu.cn
[1] School of Engineering, Sichuan Normal University, Chengdu Sichuan 610101, China

Texture is one of the most common features of vision measurement, a global feature of images, which reflects the spatial correlation of pixels [18]. Gray Level Co-occurrence Matrix is a widely used statistical method to extract texture features, which is achieved by establishing the joint probability distribution of the gray and the spatial relationship between the distance and direction of two pixels [19-21]. Although Gray Level Co-occurrence Matrix is applied to describe the spatial distribution of similar pixels of image, it is more valuable to reflect the distribution of similar edge contours [22]. In order to character the edge information, the Gray Gradient Co-occurrence Matrix establishes a joint probability distribution of gray and gradient by calculating the gradient of two pixels. Gray is usually the major parts of images and the values to the neighbors more probably than edge and noise pixels, it's local spatial properties is the key to characterize texture of the different kinds of pixels [23]. The distribution of image area has advantage on reflecting texture features, the greater the degree of face correlation is, the greater the likelihood that the pixels belong to the same texture is. Since the rank of the matrix can reflect the degree of correlation of pixels in the region, the bigger the rank of the matrix is, the smaller the correlation is, which also shows that the smaller the area of the similar texture in the region is, and otherwise the opposite.

In order to further improve the measurement accuracy of dust concentration, The Gray level-Rank Co-occurrence Matrix (GRCM) and its characteristic Moment of inertia calculation method are proposed from the global face texture feature. The Moment of inertia is used as the indirect quantity to characterize the dust concentration, and the polynomial mathematical model between the dust concentration and the Moment of inertia (PCM) is constructed to realize the dust concentration inspection.

## 2. PCM Measurement Algorithm, GRCM and Moment Of Inertia

### 2.1 PCM Measurement Algorithm

While the dust concentration $c^*$ is measured by the dust concentration calibrator, the GRCM and the Moment of inertia of corresponding dust gray image are also calculated. As shown in Table 1, the white area in the image of the GRCM represents the distribution of gray scale in the dust image.

**Tab.1 Standard dust concentrations, GRCM images and their Moments of inertia of dust images**

| $c^*$/(mg/m$^3$) | 0.17 | 9.26 | 31.73 | 98.00 | 137.00 | 187.00 | 247.00 | 275.00 | 350.00 | 466.00 | 840.00 | 949.00 |
|---|---|---|---|---|---|---|---|---|---|---|---|---|
| GRCM | 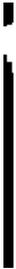 | 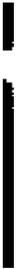 | 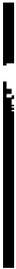 | 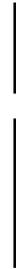 | 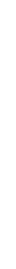 | 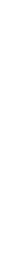 | 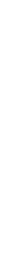 | 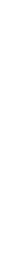 | 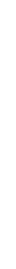 | 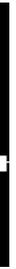 | 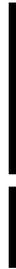 | 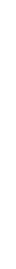 |
| Moment of inertia | 432 | 1094 | 1573 | 3153 | 3863 | 4943 | 6107 | 6916 | 8176 | 9914 | 15910 | 18560 |
| Normalization | 0.000 | 0.033 | 0.076 | 0.151 | 0.210 | 0.270 | 0.313 | 0.360 | 0.430 | 0.532 | 0.855 | 1.000 |

As is shown in Table 1, the dust concentration is positively correlated with the gray scale of dust distribution interval in GRCM, which means the higher the dust concentration is, the higher the gray scale of the distribution interval is. The dust concentration is positively correlated with the Moment of inertia of GRCM. That is, the Moment of inertia is increasingly along with the dust concentration growth. Therefore, the dust concentration can be measured by the Moment of inertia. The mathematical model between dust concentration and the Moment of inertia of GRCM is established by data fitting method, which based on the standard dust concentration measured by dust concentration calibrator and its corresponding Moment of inertia of GRCM, the expression is given as follows.

$$c(s) = k_1 s^3 + k_2 s^2 + k_3 s \quad (1)$$

where $c$ is the dust concentration, $s$ is the Moment of inertia of GRCM image of dust gray image to be measured, and $k_1$, $k_2$ and $k_3$ is the fitting coefficient which is related to the type of dust.

**2.2 GRCM**

The rank of matrix reflects the correlation between rows and cols of a matrix (region). The smaller the rank of the region is, the stronger the changing regularity of gray scale is, which shows that the local texture is similar. Other, the changing is in disorder. Define the regularized rank matrix $\mathbf{M}$ of the dust gray scale image $\mathbf{I}$, $\mathbf{M}$ is given as follows.

$$\mathbf{M}(x, y) = \mathrm{N}(\mathop{\mathrm{R}}_{w \in \Omega(x,y)}(\mathbf{I}(w))) \times L_M \quad (2)$$

where $\Omega(x, y)$ is a sub-window of size w at the centre of $x$ and $y$, $\mathrm{R}(.)$ is the rank operator, $\mathrm{N}(.)$ is the normalized operator, and $L_M$ is the rank quantization level of $\mathbf{M}$ (no greater than $W$).

The regularized matrix $\mathbf{F}$ of gray scale is processed by the normalization and quantization on the gray image $\mathbf{I}$.

$$\mathbf{F} = \mathrm{N}(\mathbf{I}) \times L_I \quad (3)$$

where $L_I$ is the quantization level of the gray scale $\mathbf{I}$ (no more than the 256-level of gray scale).

The elements $\mathbf{H}(i, j)$ in the GRCM $\mathbf{H}$ are defined as the total number of pixels that jointly satisfy the gray value $i$ and the rank value $j$ in the regularized dust gray image $\mathbf{F}$ and the regularized rank matrix $\mathbf{M}$. The element $\mathbf{H}(i, j)$ in the GRCM is given by:

$$\{\mathbf{H}(i, j) \mid \mathbf{I}(x, y) = i, \ \mathbf{G}(x, y) = j\} \quad (4)$$

where $i = 0, 1, 2, \ldots, L_I - 1$, $j = 0, 1, 2, \ldots, L_M - 1$.

At last, the probability form (P) of GRCM can be derived as follows.

$$\mathbf{P}(i, j) = \frac{\mathbf{H}(i, j)}{\sum_i \sum_j \mathbf{H}(i, j)} \quad (5)$$

**2.3 Moment of Inertia**

The Moment of inertia is the measure of inertia when the dust in GRCM rotates around the origin, which depends on the mass distribution of the dust itself and the position of the rotating shaft. The greater the dust concentration is, the greater the Moment of inertia of GRCM is, and its formula can be written as follow.

$$J_k = \sum_i \sum_j (i^2 + j^2) \times \mathbf{P}(i, j) \quad (6)$$

In the formula, $\mathbf{P}(i, j)$ represents the dust mass at $(i, j)$ as well as $i$ and $j$ respectively represent the vertical and horizontal distances from the origin, and $J_k$ represents the cumulative sum of the Moment of inertia that the dust revolves round the origin in group $k$.

In order to simplify the calculation, the moment of inertia is normalized, and the formula for calculating the normalization is as follows:

$$J = \{J_k \mid k = 1, 2, \ldots, n\} \tag{7}$$

$$s = \frac{J_k - J_{\min}}{J_{\max} - J_{\min}} \tag{8}$$

where $J$ is the set which is composed by n groups of moment of inertia $J_k$, $s$ is the normalized moment of inertia, $J_{\min}$ is the minimum value of the set $J$, $J_{\max}$ is the maximum value of the set $J$.

## 3. Experimental Process

### 3.1 Experimental Design

In order to detect dust concentration, simulate the cement production place and others, and built the dust concentration vision measurement experimental system which is shown in Fig 1.

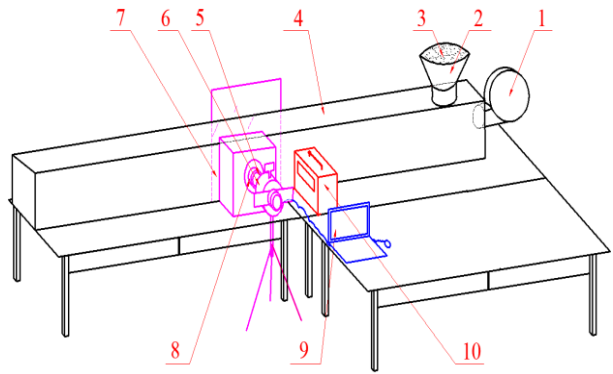

| 1-fan | 2- funnel | 3- dust | 4- pipe | 5- lens |
| 6- industrial camera | 7- black background | 8- ring light source | 9- computer | 10- CCZ-1000 |

**Fig.1 Schematic diagram of the experimental vision system**

Dust 3 leaking out from funnel 2 is blown out by Blower1. Then the dust reaches to image sampling area through pipe4. Using the industrial camera 6 with the lens 5 collects the dust image sample under the auxiliary action of the black background 7 and the ring light source 8. Next, the images are transmitted to the computer 9 to be analyzed and processed. Meanwhile, the standard of dust concentration in this current area is detected by the CCZ-1000 dust measuring instrument 10. Otherwise, the camera resolution is set to 640×360, $L_M$ =3, $L_I$ =256, coming into being a total of 21 groups of dust image samples, which 9 groups are set to be fitting samples used to fit the model and the rest are set to be test samples used to verify the model. The typical samples are shown in Fig 2.

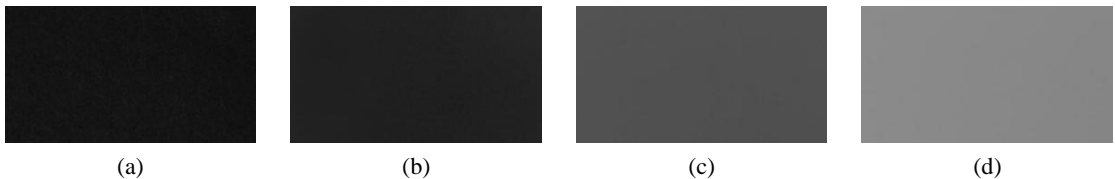

(a)      (b)      (c)      (d)

**Fig.2 Typical samples of the dust grey images at standard concentration. (a) Average of dust concentration is 0.17 mg/m³. (b) Average of dust concentration is 9.26 mg/m³. (c) Average of dust concentration is 267.00 mg/m³. (d) Average of dust concentration is 949.00 mg/m³.**

### 3.2 Selection of Window Parameter $w$

Scan the dust gray images respectively by 3×3, 5×5, 7×7, 9×9 sub-windows. At the same time, calculate the rank, regularization-rank-matrix, GRCM and Moment of inertia. Respectively, using the data fitting method to train the fitting samples to obtain the mathematical model of different window size of dust concentration and the Moment of inertia. Finally, count the measurement accuracy of the test samples. The parameters of the measurement accuracy of different window are shown in Fig 3.

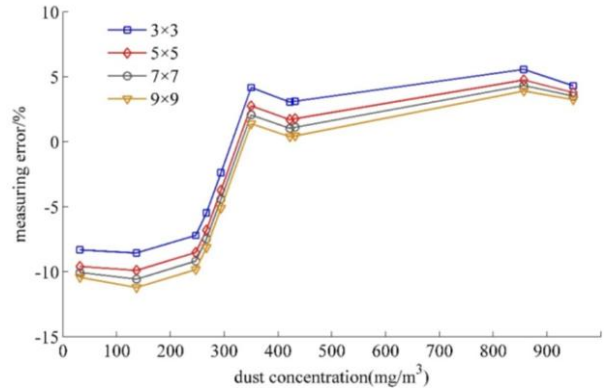

**Fig.3 Comparison of measurement accuracy of different window parameters**

It can be seen from Fig 3, the measurement accuracy of small window is higher and its measuring error is within 9% when the concentration is less than 300 mg/m$^3$. Although the measurement accuracy of the large window is higher than the small window, the measuring error of the small window also keeps within 5% when the concentration is higher than 300 mg/m$^3$. Considering the requirement for measurement accuracy of small dust concentration is relatively high in practice and the measurement speed of large window is rather slow, also with some other factors, so 3×3 window is selected to fit and test dust samples in this paper.

**3.3 PCM Model Fitting**

PCM model is obtained by fitting polynomial based on fitting samples. The fitting results are shown in following Fig 4.

$$c(s) = -1200s^3 + 1950s^2 + 240s \qquad (9)$$

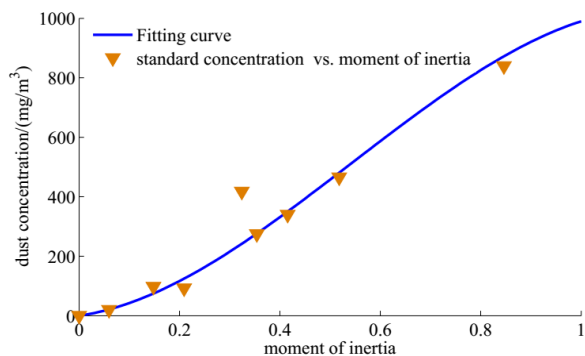

**Fig.4 Fitting results of PCM model**

It can be seen from Fig 4, it conclude that the dust concentration is directly proportional to the Moment of inertia, which can be used to measure dust concentration.

## 4. Result Analysis

The dust concentrations of test samples are calculated by PCM model and the comparisons between the testing concentration and the standard concentration are shown in table 2.

Table 2 shows that the higher the dust concentration is, the greater the Moment of inertia is. The PCM measurement error is within ±9% while the measurement range is 0.5~1000 mg/m$^3$.

**Tab.2 Measurement results of test samples**

| Sample number | Moment of inertia | Standard concentration /(mg/m$^3$) | Measured concentration/(mg/m$^3$) | Average error/% |
|---|---|---|---|---|
| 1 | 1094 | 9.26 | 10.06 | 8.65 |
| 2 | 1573 | 31.73 | 30.68 | -3.31 |
| 3 | 3863 | 137.00 | 139.39 | 2.00 |
| 4 | 4943 | 187.00 | 196.39 | 5.02 |
| 5 | 6107 | 247.00 | 260..34 | 5.40 |
| 6 | 6481 | 267.00 | 281.47 | 5.42 |
| 7 | 7021 | 294.00 | 312.37 | 6.25 |
| 8 | 8176 | 350.00 | 379.76 | 8.50 |
| 9 | 9180 | 422.00 | 439.50 | 4.15 |
| 10 | 9330 | 432.00 | 448.45 | 3.81 |
| 11 | 16412 | 857.00 | 864.33 | 0.86 |
| 12 | 18560 | 949.00 | 975.13 | 2.75 |

The comparisons of performance parameters between PCM and conventional measurement are shown in Table 3.

**Tab.3 Comparison of performance parameters between PCM measurement and traditional measurements**

| Model | Method | Measuring error | Range/(mg/m$^3$) | Period/s | Price/$ |
|---|---|---|---|---|---|
| CCZ-1000 | Filter membrane | ±10% | 0.5~1000 | 10~300 | 3889 |
| P-5FC | light scattering | ±10% | 0.01~100 | 6 | 1956 |
| LD-3F | Laser | ±10% | 0.01~100 | 1 | 1449 |
| PCM | Machine vision | ±9% | 0.5~1000 | 5 | 1391 |

1) The measurement accuracy of PCM is higher than that of traditional measurement, because it has introduced a better face texture features as the standard for calculating dust concentration.

2) PCM vision measurement range is higher than Light Scattering measurement and Laser measurement, and is equal to the Membrane Weighing measurement range. At the same time, PCM Vision measurement period is better than Membrane Weighing measurement and Light Scattering measurement, but lower than the Laser measurement. This is because it takes more time to work out the rank, but it still meets the real-time measurement requirements.

3) The cost of PCM vision system is lower than the traditional measurement system, and it has the common advantages of the vision system, which are conducive to promote.

## 5. Conclusion

On account of that the dust concentration is positively correlated with the GRCM's Moment of inertia, the dust concentration can be detected by the Moment of inertia.

When calculating the Rank matrix, the measurement accuracy of the small concentration will be higher if the small window is chosen. However, for the high concentration, the measurement accuracy shows no significant difference no matter which window has been chosen.

The dust concentration can be measured by the previously proposed PCM vision measurement and experimental system. Meanwhile, it has advantages in the measurement accuracy and the cost. The next steps: (1) to optimize the rank algorithm to reduce the operation time; (2) to verify PCM vision measurement by measuring dust concentration of other types.

## Acknowledgements

This work was supported in part by the Sichuan Science and Technology Program 2018SZ0351, the MOE (Ministry of Education in China) Project of Humanities and Social Sciences 17XJC630004, the Science Research Found of Sichuan Provincial Education Department 17ZA0329, and the Key Technology Projects for Prevention and Control of Serious Accidents in Production Safety of State Administration of Work Safety sichuan-0003-2017AQ.